\documentclass[sigconf]{acmart}

\AtBeginDocument{%
  \providecommand\BibTeX{{%
    \normalfont B\kern-0.5em{\scshape i\kern-0.25em b}\kern-0.8em\TeX}}}

\copyrightyear{2024}
\acmYear{2024}
\setcopyright{acmlicensed}\acmConference[MM '24]{Proceedings of the 32nd ACM International Conference on Multimedia}{October 28-November 1, 2024}{Melbourne, VIC, Australia}
\acmBooktitle{Proceedings of the 32nd ACM International Conference on Multimedia (MM '24), October 28-November 1, 2024, Melbourne, VIC, Australia}
\acmDOI{10.1145/3664647.3681176}
\acmISBN{979-8-4007-0686-8/24/10}

\usepackage{multirow}
\usepackage[ruled,vlined,linesnumbered]{algorithm2e}
\usepackage{setspace}

\usepackage{amssymb}
\usepackage{amsthm}
\usepackage{float}
\usepackage{bm}
\usepackage{xcolor}

\begin{document}

\title{PS-TTL: Prototype-based Soft-labels and Test-Time Learning for Few-shot Object Detection}

\author{Yingjie Gao}
\orcid{0009-0003-4037-4701}
\affiliation{
  \institution{SKLCCSE, Beihang University}
  \city{Beijing}
  \country{China}}
\email{gaoyingjie@buaa.edu.cn}

\author{Yanan Zhang}
\orcid{0000-0003-1592-1067}
\affiliation{
  \institution{SKLCCSE, Beihang University}
  \city{Beijing}
  \country{China}}
\email{zhangyanan@buaa.edu.cn}

\author{Ziyue Huang}
\orcid{0009-0008-8452-5539}
\affiliation{
  \institution{SKLCCSE, Beihang University}
  \city{Beijing}
  \country{China}}
\email{ziyuehuang@buaa.edu.cn}

\author{Nanqing Liu}
\orcid{0000-0001-7564-4896}
\affiliation{
  \institution{SIST, Southwest Jiaotong University}
  \city{Chengdu}
  \country{China}}
\email{lansing@my.swjtu.edu.cn}

\author{Di Huang}
\orcid{0000-0002-2412-9330}
\authornote{Corresponding Author.}
\affiliation{
  \institution{SKLCCSE, Beihang University}
  \city{Beijing}
  \country{China}
  \institution{\\Zhejiang Industrial Big Data and Robot Intelligent System Key Lab.}
  \city{Hangzhou}
  \country{China}}
\email{dhuang@buaa.edu.cn}

\renewcommand{\shortauthors}{Yingjie Gao et al.}

\begin{abstract}
  In recent years, Few-Shot Object Detection (FSOD) has gained widespread attention and made significant progress due to its ability to build models with a good generalization power using extremely limited annotated data. The fine-tuning based paradigm is currently dominating this field, where detectors are initially pre-trained on base classes with sufficient samples and then fine-tuned on novel ones with few samples, but the scarcity of labeled samples of novel classes greatly interferes precisely fitting their data distribution, thus hampering the performance. To address this issue, we propose a new framework for FSOD, namely Prototype-based Soft-labels and Test-Time Learning (PS-TTL). Specifically, we design a Test-Time Learning (TTL) module that employs a mean-teacher network for self-training to discover novel instances from test data, allowing detectors to learn better representations and classifiers for novel classes.
  Furthermore, we notice that even though relatively low-confidence pseudo-labels exhibit classification confusion, they still tend to recall foreground. We thus develop a Prototype-based Soft-labels (PS) strategy through assessing similarities between low-confidence pseudo-labels and category prototypes as soft-labels to unleash their potential, which substantially mitigates the constraints posed by few-shot samples. Extensive experiments on both the VOC and COCO benchmarks show that PS-TTL achieves the state-of-the-art, highlighting its effectiveness. The code and model are available at \url{https://github.com/gaoyingjay/PS-TTL}.
\end{abstract}



\maketitle

\section{Introduction}
 Object detection~\cite{zhang2021pc,yolov7,faster,zhou2023octr} is a fundamental task of computer vision and multimedia, involving a variety of applications, including autonomous driving~\cite{zhang2023sa,zhang2022cat}, robotic manipulation~\cite{ma2023towards,qin2023rgb}, medical analysis~\cite{liu2019privacy,li2019clu}, \emph{etc}. Despite that significant progress has been achieved in recent years~\cite{fcos,atss,rank_detr,dino}, detectors heavily rely on many training samples. Considering that labeling data is rather expensive and collecting examples for rare categories is extremely hard, solutions are required to deal with data-limited scenarios.
 
 Few-Shot Object Detection (FSOD) is a promising way to address this issue. It aims to train an object detector using only a few samples on novel classes with the help of abundant data on base classes, which has received widespread attention from both the academia and industry. Early FSOD methods typically adopt the meta-learning paradigm, organizing object detection into a series of episode tasks with few-shot samples, where each episode includes a support set of $N$-way $K$-shot images and a query set. The support set is utilized for model training with a limited number of samples, while the query set is employed to assess the performance of the model on novel objects. Kang \emph{et al.}~\cite{kang2019few} propose a lightweight feature reweighting module that learns to capture global features of support images and embeds such features into reweighting coefficients to adjust meta features of query images. Meta R-CNN~\cite{yan2019meta} conducts meta-learning on Reigion-of-Interest (RoI) features instead of those of full images to better fit the detection problem. Subsequent alternatives progress from optimizing both classification and localization features~\cite{wang2019meta,fan2020few,han2021query,yang2020restoring,demirel2023meta,liu2023transformation}, by employing Transformer to capture spatial relationships between support and query classes~\cite{han2022few,zhang2022meta} as well as exploring inter-class relationships~\cite{zhang2021accurate,karlinsky2019repmet,han2023few,lu2023breaking}. However, such methods suffer complex architectures and training procedures with increased computational costs. Additionally, they are criticized for its interpretability of what the model learns in the novel stage.
 
\begin{figure}[t]
  \centering
  \includegraphics[width=0.97\linewidth]{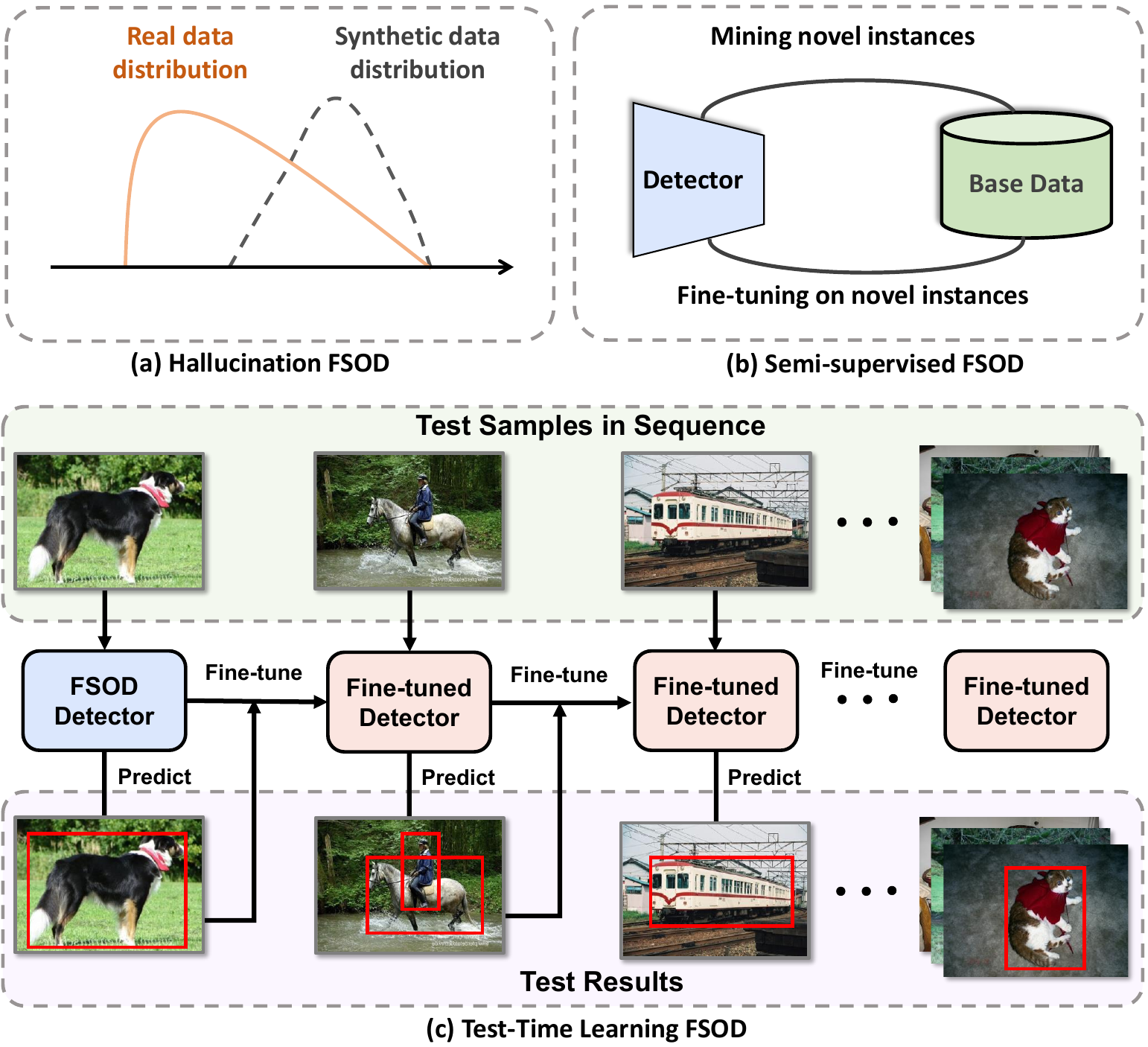}
  \caption{Motivation of FSOD with Test-Time Learning. (a) Hallucination methods suffer from the distribution gap between synthetic and real data. (b) Semi-supervised methods mine implicit novel instances from base data; however, novel instances do not always appear in base data. (c) For the first time, we propose to learn an enhanced model at test-time, effectively leveraging data of novel classes present in test data in a more realistic manner aligned with real-world applications.}
  \label{fig:fig1}
\end{figure} 

 To achieve fast training and simple deployment for adaptation to novel classes, most existing FSOD methods follow a fine-tuning based paradigm. The detector is first pre-trained on base classes with adequate samples and then fine-tuned on novel ones with few samples. Early attempts~\cite{chen2018lstd,wu2020multi} employ a jointly fine-tuning based architecture, where the entire pre-trained base model, comprising both the class-agnostic and class-specific layers, is simultaneously updated during training on the novel task. Later, the two-stage fine-tuning methods~\cite{wang2020frustratingly,sun2021fsce,cao2021few,fan2022few,wu2021universal,zhang2020cooperating,ma2022few} demonstrate that maintaining the feature extraction part of the model unchanged and solely fine-tuning the last layer can significantly boost the accuracy. Based on this fact, most of the subsequent methods are combined with knowledge distillation~\cite{wu2022multi,pei2022few,nguyen2022few}, context reasoning~\cite{zhu2021semantic,kim2021spatial}, or decoupling detection networks~\cite{qiao2021defrcn,yang2022efficient,lu2022decoupled} to further improve the detection performance. Unfortunately, constrained by the limited samples of novel classes, they struggle to precisely capture the data distribution. Some methods address this issue by generating synthetic data for novel classes~\cite{zhang2021hallucination,zhao2022exploring} or mining implicit novel instances from the training set~\cite{kaul2022label,cao2022mini,tang2023semi}. But the former synthesize novel samples according to the base data and the results often deviate from the true distribution as shown in Fig.~\ref{fig:fig1}(a), while the latter rely on the assumption that unlabeled novel instances are widely present in abundant base data as shown in Fig.~\ref{fig:fig1}(b), which may not always hold in real-world cases.
  
 The accessibility of novel instances in test data motivates us to explore a new framework which fine-tunes an object detection model at test-time as shown in Fig.~\ref{fig:fig1}(c). Compared to mining novel instances from base data (the presence of unlabeled novel instances in data of base classes essentially incurs a loophole in FSOD settings), conducting online learning on test data is a more realistic manner aligned with real-world applications. In this paper, we propose a Test-Time Learning (TTL) module, which utilizes a mean-teacher network for self-training to simultaneously train and infer on test data, effectively leveraging novel instances present in test data. Specifically, both the student and teacher networks are first initialized by the few-shot detector fine-tuned on novel data. Then, the teacher network takes test data as input to generate pseudo-labels. The student model is trained using these pseudo-labels after post-processing and $N$-way $K$-shot data as supervision signals and updates the teacher through Exponential Moving Average (EMA). Additionally, considering the limited number of high-quality pseudo-labels and the fact that a large number of low-quality pseudo-labels can recall foreground but exhibit a low classification accuracy, we develop a Prototype-based Soft-labels (PS) strategy to unlock the potential of these low-quality pseudo-labels. In this case, we maintain class prototypes and compute the feature similarity between low-confidence pseudo-labels and class prototypes to replace them with soft-labels. Class prototypes are initialized using $N$-way $K$-shot data and dynamically updated during online learning using instance features of high-confidence pseudo-labels. Finally, we integrate the two modules aforementioned to build a new framework for FSOD, dubbed PS-TTL.

 In summary, the major contributions of this paper includes:
 \begin{itemize}
 \item We propose a novel PS-TTL framework for FSOD, which effectively mines implicit novel instances from test data to address the issue of limited samples of novel classes. To the best of our knowledge, it is the first attempt to explore fitting data distribution of novel classes in a way that is more in line with real-world scenarios. 
 \item We design the TTL module that adopts a mean-teacher network for self-training to discover novel instances on test data and develop the PS strategy to unleash the power of low-quality pseudo-labels.
 \item We achieve a newly state-of-the-art performance of all few-shot settings on the VOC and COCO benchmarks in comparison to the published counterparts, demonstrating its advantage in detecting novel objects.
\end{itemize}

\section{Related Work}
\subsection{Object Detection}
Object detection aims to identify and localize objects within images, constituting a fundamental challenge in the fields of computer vision and multimedia. Recently, the success of deep learning has triggered numerous effective object detection methods. They can be roughly categorized into two main groups: single-stage and two-stage. Single-stage detectors (\emph{e.g.} SSD \cite{liu2016ssd} and RetinaNet \cite{retinanet}) predict bounding boxes and classification scores based on pre-defined anchors, exhibiting good real-time performance. 
The YOLO \cite{redmon2016you, yolov7} series, by continuously assimilating the latest advancements, such as label assignment and multi-scale feature fusion, has delivered high-precision real-time object detection. Although the structure of single-stage detectors is straightforward and efficient, their integrated design makes them less adaptable to FSOD tasks. In contrast, two-stage detectors (\emph{e.g.} Faster R-CNN \cite{faster}) usually first use a Region Proposal Network (RPN) to generate potential proposals and then refine them by other modules. Compared to single-stage detectors, two-stage ones report higher detection performance and are commonly used for FSOD.

\vspace{-0.2cm}
\subsection{Few-shot Object Detection}
FSOD methods enable detectors to rapidly adapt to new objects with minimal data while preserving competitive performance. There are mainly two paradigms: meta-learning based and fine-tuning based. Meta-learning based methods \cite{wang2019meta,li2021beyond,li2021transformation,yin2022sylph} employ a series of $N$-way $K$-shot detection tasks for training, aiming to well generalize to novel tasks with limited samples. FSRW \cite{kang2019few} proposes a feature reweighting strategy, which extracts class-specific representations from support images and utilizes them to reweight the importance of query features. Similarly, Meta R-CNN \cite{yan2019meta} combines a two-stage detector and reweights RoI features in the detection head. Attention-RPN \cite{fan2020few} exploits matching relationship between the few-shot support set and query set with a contrastive training scheme, which can then be applied to detect novel objects without retraining and fine-tuning. QA-FewDet \cite{han2021query} uses a graph model to capture multi-relations among the proposal and class nodes. While theoretical promising, their training and inference processes are highly complex, making them difficult to deploy in real-world scenarios.

Fine-tuning based methods leverage a two-stage training procedure, \emph{i.e.}, first base training and then few-shot fine-tuning, which expects to transfer prior knowledge from base classes to novel ones. LSTD~\cite{chen2018lstd} is the earliest attempt in this domain, incorporating detection knowledge transfer and background depression regularization. 
TFA \cite{wang2020frustratingly} simply freezes the backbone and only fine-tunes the detection head with novel classes, which largely improves the performance. Subsequent research refines the TFA method and integrates it with other techniques to further enhance the FSOD performance \cite{sun2021fsce,nguyen2022few, pei2022few, wu2022multi, kim2021spatial, zhu2021semantic, lu2022decoupled, qiao2021defrcn, fan2021generalized, yang2022efficient}. FSCE \cite{sun2021fsce} introduces contrastive learning to learn discriminative object proposal representations, alleviating the misclassification issue in novel classes. DeFRCN \cite{qiao2021defrcn} employs the gradient decoupled layer for multi-stage decoupling and the prototypical calibration block to
align original classification scores. Although fine-tuning based methods demonstrate satisfactory results, the limited samples of novel classes still make it challenging for detectors to capture accurate data distributions.

To alleviate the issue above, HallucFsDet \cite{zhang2021hallucination} introduces a hallucinator network trained on base classes to synthesize additional training examples for novel classes. Zhao \emph{et al.} \cite{zhao2022exploring} assume that the features of both base and novel classes follow a Gaussian distribution and generate samples for novel classes according to the variance of similar base classes. However, these hallucination methods relying on base classes tend to result in biased synthetic novel samples. The other semi-supervised FSOD methods \cite{cao2022mini, kaul2022label, tang2023semi} investigate mining implicit novel instances from training data with the assumption that novel instances do appear in base data. Kaul \emph{et al.} \cite{kaul2022label} present a simple pseudo-labelling strategy to detect potential novel instances in base datasets while MINI \cite{cao2022mini} comprises an offline and online mechanism, facilitating novel instance mining. However, as such an assumption does not always hold (\emph{e.g.} in rare disease or animal detection), they are problematic in practical cases. Considering the accessibility of novel instances in test data, we are motivated to explore fine-tuning detectors at test-time.

\section{Method}

\subsection{Problem Formulation}
\label{sec:Problem Setting}
We follow the standard FSOD setting introduced in~\cite{wang2020frustratingly} with two disjoint training sets: a base dataset $D^b=\{x^b_i,y^b_i\}$ with exhaustively annotated instances for each base class $C^b$ and a novel dataset $D^n=\{x^n_i,y^n_i\}$ with only $K$ instances (usually less than 30) for each novel class $C^n$, where $x_i$ and $y_i$ refer to the input image and the Ground Truth (GT) label, respectively. It is worth noting that there is no intersection between base and novel classes, \emph{i.e.}, $C^b \cap C^n = \emptyset$. In this case, the ultimate goal of FSOD is to train a robust detector based on $D^b$ and $D^n$ to deal with the objects in the test set that contains both types of instances in $C^b \cup C^n$.

\begin{figure*}[t]
  \centering
  \includegraphics[width=0.95\textwidth]{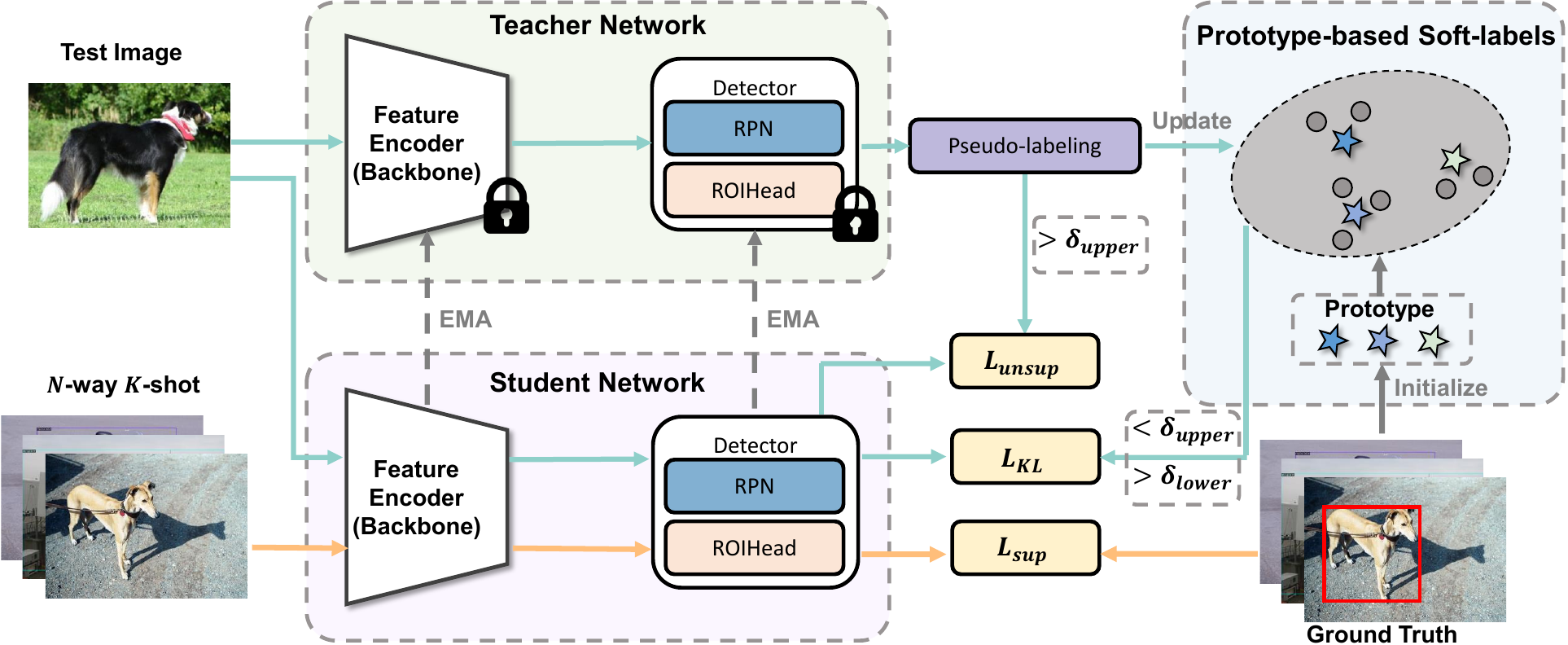}
  \vspace{-0.2cm}
  \caption{The overview of the proposed Prototype-based Soft-labels and Test-Time Learning (PS-TTL) framework for FSOD. Both the student and teacher networks are first initialized by the few-shot detector and then fine-tuned on test data. The teacher network takes test data as input to generate pseudo-labels, while the student model is trained using these pseudo-labels after post-processing with $N$-way $K$-shot data as supervision signals and updates the teacher network through EMA. A Prototype-based Soft-labels (PS) strategy is adopted to maintain class prototypes and compute the feature similarity between low-confidence pseudo-labels and class prototypes to replace them with soft-labels.}
  \label{fig:overall}
\end{figure*}

\subsection{Base Detector}
\label{sec:DeFRCN}
DeFRCN \cite{qiao2021defrcn} is a state-of-the-art fine-tuning based few-shot object detector, built through two training stages. In the first phase, Faster-RCNN is trained on base classes $C^b$ with sufficient samples, while in the second phase, transfer learning is conducted by fine-tuning Faster-RCNN on base classes and novel classes $C^b \cup C^n$ with $K$ instances per class. Fine-tuning on a balanced set $D^{balanced}$ containing training samples for both base and novel classes helps to preserve the performance on base classes. The entire procedure is summarized as follows:
 \begin{equation}
  M_{init} \xrightarrow[]{D^b} M_{base} \xrightarrow[]{D^{balanced}} M_{novel}
 \end{equation}
where $M_{init}$, $M_{base}$, and $M_{novel}$ indicate the detector in the initialization, base training, and novel fine-tuning stages, respectively.
 
Different from previous fine-tuning based methods, which only fine-tune a small number of parameters of Faster-RCNN, such as the prediction head, to avoid overfitting of the detector, DeFRCN introduces a gradient decoupled layer during fine-tuning to stop the gradient between RPN and backbone while scaling the gradient between RCNN and backbone. This allows the detector to sufficiently learn from novel data while preventing overfitting, making DeFRCN remarkably superior to other existing counterparts. 

Despite the significant progress made by the fine-tuning based methods, given only $K$ novel instances, they fail to accurately capture the data distribution. To overcome this obstacle, we propose Prototype-based Soft-labels and Test-Time Learning (PS-TTL) to mine novel instances in test data. The overall architecture of the model is illustrated in Fig. \ref{fig:overall}.

\subsection{Test-Time Learning with Mean-Teacher}
\label{sec:ttt}
 Self-training has shown promising performance for semi-supervised object detection \cite{lee2013pseudo, tarvainen2017mean, liu2021unbiased}. It typically predicts pseudo-labels for unlabeled data, where high-confidence pseudo-labels are used to supervise detector training. 
 
 In this work, we aim to fully leverage novel instances in test data, especially in the scenario of online learning, called Test-Time Learning (TTL). To this end, we employ a mean-teacher self-training paradigm \cite{tarvainen2017mean}, which mainly consists of two architecturally identical detectors, \emph{i.e.} the student network and the teacher network. The teacher network first works on test data and renders pseudo-labels from its detection results through post-processing procedures (\emph{e.g.}, Non-Maximum Suppression (NMS) and filtering using a confidence threshold). The high-quality pseudo-labels are screened out to supervise the student network, enhancing its capability. 
 
 Since the teacher network is expected to produce reliable pseudo-labels of novel classes in test data, the few-shot detector $M_{novel}$ which has been fine-tuned on novel data is used to initialize both the student and teacher networks. However, the self-training paradigm inevitably generates noisy pseudo-labels, particularly for novel classes. Considering that excessively noisy pseudo-labels deteriorate the performance of the student network as training progresses, to address this issue, we first apply NMS for each class to remove duplicate detection boxes. Then, we set a high confidence threshold $\delta_{upper}$ to exclude uncertain labels. Finally, we optimize the student network using the remaining high-quality pseudo-labels with the loss function as follows:
 \begin{equation}
  L_{unsup}(X^t, \hat{Y^t}) = L^{rpn}_{cls}(X^t, \hat{Y^t}) + L^{roi}_{cls}(X^t, \hat{Y^t})
  \label{equation:eq2}
 \end{equation}
 where $X^t$ is the input test image and $\hat{Y^t}$ denotes the filtered pseudo-label. Note that the unsupervised loss is only applied to the classification heads of RPN and RoI.

 On the other side, as the few-shot detector $M_{novel}$ is not strong enough, pseudo-labels still contain noise even after filtering out low-confidence predictions. Therefore, to alleviate the degradation of the few-shot detector during test-time learning, we propose to take $N$-way $K$-shot data $D^{balanced}$ as supervision signals. The supervised loss for training the student network can thus be defined as:
 \begin{align}
  L_{sup}(X^s,Y^s) &= L^{rpn}_{cls}(X^s, {Y^s}) + L^{rpn}_{reg}(X^s, {Y^s}) \notag \\ 
  &+ L^{roi}_{cls}(X^s, {Y^s}) + L^{roi}_{reg}(X^s, {Y^s}) 
  \label{equation:eq3}
 \end{align}
 where $\{X^s,Y^s\} \in D^{balanced}$. Both the RPN and RoI heads adopt the classification loss and bounding box regression loss.
 
 Following the mean-teacher framework~\cite{tarvainen2017mean}, we update the weights of the student network using both $L_{unsup}$ and $L_{sup}$. To obtain high-quality pseudo-labels from test data, we update the weights of the teacher network via Exponential Moving Average (EMA) of those of the student as below:
 \begin{equation}
   \theta_t = \alpha \theta_t + (1 - \alpha) \theta_s
 \end{equation}
where $\theta_t$ and $\theta_s$ are the network parameters of the teacher and the student, respectively. $\alpha$ is the EMA momentum coefficient.

\subsection{Prototype-based Soft-labels}
\label{sec:psl}
 The mean-teacher self-training framework presented in Section \ref{sec:ttt} for TTL on test data promotes the detection performance, and a large threshold $\delta_{upper}$ to filter the generated pseudo-labels is more beneficial (please see the experiments). However, this leads to severe detection missing, and many test images do not have any pseudo-label. Different from semi-supervised object detection, where multiple rounds of fine-tuning can be conducted on unlabeled data, under the TTL setting, we can only perform one epoch of training on test data, making it challenging to fully utilize every input test image.

\begin{figure}[t]
  \centering
  \includegraphics[width=0.98\linewidth]{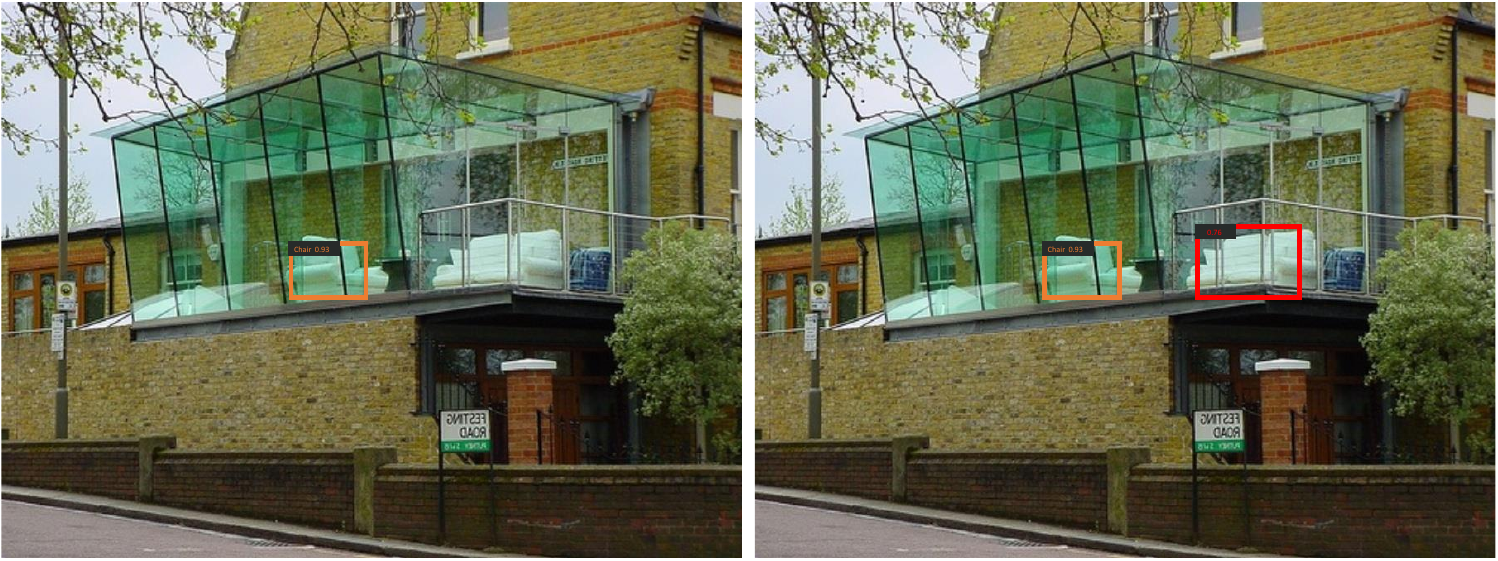}
  \vspace{-0.2cm}
  \caption{Illustration of the issue of detection missing. In the left image, many pseudo-labels are filtered by $\delta_{upper}$ and most objects are not detected. In the right image, when $\delta_{lower}$ is applied, some relatively low-confidence pseudo-labels are retained as high-quality implicit foreground predictions.}
  \label{fig:problem}
\end{figure}

 As shown in Fig. \ref{fig:problem}, we observe that relatively low-confidence pseudo-labels, despite suffering from classification confusion, mostly recall foreground. Based on this phenomenon, we propose a Prototype-based Soft-labels (PS) strategy to replace the hard labels of these implicit foreground predictions with soft-labels for fully unleashing the potential of low-quality pseudo-labels.

 Specifically, we first introduce a lower bound confidence threshold $\delta_{lower}$, and the predicted results between $\delta_{lower}$ and $\delta_{upper}$ are also assigned as foreground. Due to unavoidable classification confusion in these implicit foreground predictions, it fails to effectively remove redundant boxes by employing class-specific NMS in the teacher network. Instead, after removing the hard labels of these implicit foreground predictions, we apply class-agnostic NMS to them using every high-confidence pseudo prediction (\emph{i.e.}, whose confidence score is greater than $\delta_{upper}$) to filter the redundant ones.
 
 We then generate soft-labels for these implicit foreground predictions by measuring their similarities to each class. Formally, given an implicit foreground prediction $r$, we define its similarity to a class $c$ as the cosine distance between its RoI feature $F_r$ and the prototype $P^c$ of the class $c$:
 \begin{equation}
   s^c_r = \frac{F^T_r P^c}{\lvert\rvert F^T_r \lvert\rvert {~} \lvert\rvert P^c \lvert\rvert}, \quad c \in C^b \cup C^n
   \label{equation:eq5}
 \end{equation}
 
 Finally, $s_r = [s^1_r,s^2_r,...,s^N_r]$ follows a softmax function to generate $u_r$, which represents the soft-label of the implicit foreground prediction $r$. We minimize the Kullback-Leibler (KL) divergence between the soft-label and the class probability of each implicit foreground prediction $r$:
 \begin{equation}
   L_{KL} = \sum_{c=1} ^ {N+1} u^c_rlog(\frac{u^c_r}{v^c_r})
   \label{equation:eq6}
 \end{equation}
 where $v_r$ is the class probability, $N+1$ denotes $N$ foreground classes and one background class. Hence, we set $u^{N+1}_r=0$.
 
 To leverage soft-labels of implicit foreground predictions at the early stage during TTL, we initialize the class prototypes with $N$-way $K$-shot data:
 \begin{equation}
   {P}^c = \frac{1}{K} \sum_{i=1}^{K} {F}^c_i
   \label{equation:eq7}
 \end{equation}
 where $F^c_i$ is the RoI feature of the $i$-th instance for class $c$. Because the $N$-way $K$-shot data cannot accurately represent the class prototypes, we propose to dynamically update them using both the supervised $N$-way $K$-shot data and test data with high-confidence pseudo-labels, enabling the class prototypes to converge to the true representations as training progresses. In this case, we update the class prototypes using the following formula:
 \begin{equation}
   P^c_{new} = P^c_{old} (1 - sim(P^c_{old}, {F^c_{avg}})) + {F^c_{avg}} sim(P^c_{old}, {F^c_{avg}})
   \label{equation:eq8}
 \end{equation}
 where ${F^c_{avg}}$ is the averaged RoI feature of high-confidence predictions for class $c$ during TTL, and $sim(\cdot, \cdot)$ denotes the normalized cosine similarity function as:
 \begin{equation}
 sim(x_1, x_2) = ( \frac{x^T_1 x_2}{\lvert\rvert x^T_1 \lvert\rvert {~} \lvert\rvert x_2 \lvert\rvert} + 1) / 2.
 \label{equation:eq8plus}
 \end{equation}

\begin{algorithm}[t]
\setstretch{1.1}
\caption{The PS-TTL algorithm.}
\label{algorithm:algo_1}

\KwIn {Test data ${D}^t=\{x^t_i\}_{i=1 \cdots N_t}$.\\
\qquad \quad $N$-way $K$-shot data ${D}^{balanced}=\{x^s_i, y^s_i\}_{i=1 \cdots NK}$.\\
\qquad \quad Few-shot detector $M_{novel}$.}
\KwOut {An improved detector.}

{Init $\text{StudentNet}, \text{TeacherNet} = M_{novel}$; init class prototypes $P^c$ according to Eq.~\ref{equation:eq7}}

\For{$i \leftarrow 1$ \KwTo $N_{t}$}{
     \textcolor{gray}{\# Inference Stage:}\\
     $r = \text{TeacherNet}(x^t_i)$\\
     
     \textcolor{gray}{\# Fine-tuning Stage:}\\
     Calculate $L_{unsup}$ by Eq.~\ref{equation:eq2} and  $L_{sup}$ by Eq.~\ref{equation:eq3}\\
     Update class prototypes $P^c$ by Eq.~\ref{equation:eq8} \\
     
     Generate soft-labels $u^c$ by Eq.~\ref{equation:eq5}; calculate $L_{KL}$ by Eq.~\ref{equation:eq6} \\
     
     Calculate $L_{total}$ by Eq.~\ref{equation:eq9}\\
     
     StudentNet update (SGD): $\theta_s = \theta_s - \eta \nabla {L}_{\text{total}}$\\
     TeacherNet update (EMA): $\theta_t = \alpha\theta_t + (1-\alpha) \theta_s$\\
     }
\end{algorithm}

\begin{table*}[ht]
\renewcommand\arraystretch{1}
  \centering
  \caption{Comparison of different FSOD methods in terms of nAP50 (\%) on three PASCAL VOC novel split sets. 
  $^\ddag$ indicates that it mines implicit novel instances from base data. }
  \resizebox{0.95\textwidth}{!}{  
  \begin{tabular}{l| c c c c c | c c c c c | c c c c c}
    \toprule
    \multirow{2}{*}{Method / Shots}  & \multicolumn{5}{c|}{Novel Split 1} & \multicolumn{5}{c|}{Novel Split 2} & \multicolumn{5}{c}{Novel Split 3} \\
    & 1 & 2 & 3 & 5 & 10 & 1 & 2 & 3 & 5 & 10& 1 & 2 & 3 & 5 & 10 \\
    \midrule
    LSTD~\cite{chen2018lstd}      &8.2&1.0&12.4&29.1&38.5&11.4&3.8&5.0&15.7&31.0&12.6&8.5&15.0&27.3&36.3\\
    FSRW~\cite{kang2019few}   &14.8&15.5&26.7&33.9&47.2&15.7&15.3&22.7&30.1&40.5&21.3&25.6&28.4&42.8&45.9   \\
    MetaDet~\cite{wang2019meta} &18.9&20.6&30.2&36.8&49.6&21.8&23.1&27.8&31.7&43.0&20.6&23.9&29.4&43.9&44.1   \\
    Meta R-CNN~\cite{yan2019meta}  &19.9&25.5&35.0&45.7&51.5&10.4&19.4&29.6&34.8&45.4&14.3&18.2&27.5&41.2&48.1 \\
    TFA w/cos~\cite{wang2020frustratingly} &39.8&36.1&44.7&55.7&56.0&23.5&26.9&34.1&35.1&39.1&30.8&34.8&42.8&49.5&49.8 \\
    MPSR~\cite{wu2020multi}      &41.7&$-$&51.4&55.2&61.8&24.4&$-$&39.2&39.9&47.8&35.6&$-$&42.3&48.0&49.7\\
    HallucFsDet~\cite{zhang2021hallucination}  &47.0&44.9&46.5&54.7&54.7&26.3&31.8&37.4&37.4&41.2&40.4&42.1&43.3&51.4&49.6\\
    Retentive R-CNN\cite{fan2021generalized} &42.4&45.8&45.9&53.7&56.1&21.7&27.8&35.2&37.0&40.3&30.2&37.6&43.0&49.7&50.1\\
    FSCE~\cite{sun2021fsce}  &44.2&43.8&51.4&61.9&63.4&27.3&29.5&43.5&44.2&50.2&37.2&41.9&47.5&54.6&58.5\\
    UP-FSOD~\cite{wu2021universal} &43.8&47.8&50.3&55.4&61.7&31.2&30.5&41.2&42.2&48.3&35.5&39.7&43.9&50.6&53.3\\
    SRR-FSD~\cite{zhu2021semantic}  &47.8&50.5&51.3&55.2&56.8&32.5&35.3&39.1&40.8&43.8&40.1&41.5&44.3&46.9&46.4\\
    DCNet~\cite{hu2021dense}  &33.9&37.4&43.7&51.1&59.6&23.2&24.8&30.6&36.7&46.6&32.3&34.9&39.7&42.6&50.7\\
    Meta FRCN~\cite{han2022meta}  &43.0&54.5&60.6&66.1&65.4&27.7&35.5&46.1&47.8&\textbf{51.4}&40.6&46.4&53.4&59.9&58.6\\
    QA-FewDet~\cite{han2021query}&42.4&51.9&55.7&62.6&63.4&25.9&37.8&46.6&48.9&51.1&35.2&42.9&47.8&54.8&53.5\\
    CME~\cite{li2021beyond} &41.5&47.5&50.4&58.2&60.9&27.2&30.2&41.4&42.5&46.8&34.3&39.6&45.1&48.3&51.5\\
    FADI~\cite{cao2021few}&50.3&54.8&54.2&59.3&63.2&30.6&35.0&40.3&42.8&48.0&45.7&49.7&49.1&55.0&59.6\\
    LVC$^\ddag$~\cite{kaul2022label}&54.5&53.2&58.8&63.2&65.7&32.8&29.2&50.7&49.8&50.6&48.4&52.7&55.0&59.6&59.6\\
    DeFRCN*~\cite{qiao2021defrcn}&55.4&62.1&65.0&68.4&67.6&35.5&45.4&51.8&51.7&47.5&50.8&57.4&57.8&62.7&\textbf{65.0}\\
    \midrule
    Ours&\textbf{58.4}&\textbf{65.7}&\textbf{67.9}&\textbf{69.3}&\textbf{68.1}&\textbf{38.4}&\textbf{47.8}&\textbf{52.8}&\textbf{53.6}&49.1&\textbf{53.0}&\textbf{58.8}&\textbf{59.2}&\textbf{63.8}&64.1\\
    \bottomrule
  \end{tabular}
  }
  \label{tab:voc}
\end{table*}

\begin{table}[t]
\renewcommand\arraystretch{1}
  \centering
  \caption{Performance (\%) of FSOD methods on MS COCO.}
\resizebox{0.45\textwidth}{!}{
  \begin{tabular}{l| c c | c c}
    \toprule
    \multirow{2}{*}{Method}  &  \multicolumn{2}{c|}{10-shot} & \multicolumn{2}{c}{30-shot}\\
    & nAP & nAP75 & nAP & nAP75\\
    \midrule
    FSRW~\cite{kang2019few}& 5.6 & 4.6&9.1 &7.6 \\
    MetaDet~\cite{wang2019meta}&7.1 &6.1 &11.3 & 8.1\\
    Meta R-CNN~\cite{yan2019meta}&8.7 & 6.6& 12.4 &10.8 \\
    TFA w/cos~\cite{wang2020frustratingly}&10.0 & 9.3&13.7  & 13.4\\
    MPSR~\cite{wu2020multi}& 9.8& 9.7& 14.1&14.2 \\
    Retentive R-CNN~\cite{fan2021generalized}&10.5 &$-$ &13.8 &$-$ \\
    FSCE~\cite{sun2021fsce}&11.9 &10.5 &16.4 &16.2 \\
    UP-FSOD~\cite{wu2021universal}&11.0 &10.7 &15.6  & 15.7\\
    SRR-FSD~\cite{zhu2021semantic}&11.3 & 9.8 &14.7 &13.5 \\
    DCNet~\cite{hu2021dense}&12.8 & 11.2&18.6 & 17.5\\
    Meta FRCN~\cite{han2022meta}& 12.7& 10.8 &16.6 & 15.8 \\
    QA-FewDet~\cite{han2021query}& 11.6 & 9.8 &16.5 &15.5 \\
    CME~\cite{li2021beyond}& 15.1 & 16.4& 16.9&17.8 \\
    FADI~\cite{cao2021few}&12.2 & 11.9 &16.1 &15.8 \\
    LVC$^\ddag$~\cite{kaul2022label}&17.8 &17.8  & 24.5&25.0 \\
    DeFRCN*~\cite{qiao2021defrcn}& 17.1 &15.9 &20.2 & 19.5 \\
    \midrule
    Ours& \textbf{17.3} &\textbf{16.7} &\textbf{20.9} &\textbf{21.3 }\textbf{}\\
    \bottomrule
  \end{tabular}
  }
  \label{tab:coco}
\end{table}

 \subsection{Training Procedure}
 \label{sec:Training Procedure}
 During TTL, the total loss to optimize is as follows:
 \begin{equation}
   L_{total}=L_{sup} + \lambda_1 L_{unsup} + \lambda_2 L_{KL}
   \label{equation:eq9}
 \end{equation}
 consisting of the supervised loss of $N$-way $K$-shot data, the unsupervised loss of pseudo-labels on test data, and the KL loss of soft-labels on test data. Here, $\lambda_1$ and $\lambda_2$ are hyper-parameters to balance such losses.  
 
 During this procedure, few-shot detectors are able to learn from test data. When a mini-batch of test samples arrive, we update the weights of the model through the total loss $L_{total}$. A detailed description is provided in Algorithm~\ref{algorithm:algo_1}.

\section{Experiments}
 
\subsection{Datasets}
\label{sec:datasets}
 \textbf{PASCAL VOC.} For PASCAL VOC \cite{everingham2015pascal}, the overall 20 classes are divided into 15 base classes and 5 novel classes. Following TFA \cite{wang2020frustratingly}, we utilize three different class splits, namely split 1, 2, and 3. For each split, base classes are exhaustively annotated, but novel classes only have $K=1,2,3,5,10$ annotated instances per class. Both base and novel instances are sampled from the PASCAL VOC (07+12) \emph{trainval} set, and the model is tested on the PASCAL VOC07 \emph{test} set. We report AP50 for novel classes during evaluation.

 \textbf{MS COCO.} MS COCO \cite{lin2014microsoft} has 80 classes, and we select the 20 classes that overlapped with PASCAL VOC as novel classes and the remaining 60 classes as base classes. In this case, we evaluate our method with $K=10,30$ shots for each novel class. We report mAP and AP75, respectively.

\subsection{Implementation Details}
\label{sec:details}
 Our method can be combined with the majority of fine-tuning based few-shot object detectors. For simplicity, we choose the most representative SOTA method, \emph{i.e.} DeFRCN \cite{qiao2021defrcn}, as our baseline. DeFRCN adopts Faster-RCNN \cite{faster} as the detection model and uses the ResNet-101 backbone pre-trained on ImageNet. We use DeFRCN, which has been pre-trained on base classes and fine-tuned on novel classes, to initialize our model, and then fine-tune it on test data. During TTL, we fine-tune our model with a mini-batch of 2 on a single GPU, which simulates the real inference process of the few-shot detector. Besides, we follow a one-epoch setting, where we fine-tune on test data for only one epoch. We also utilize the $N$-way $K$-shot data for novel fine-tuning during the testing process. Due to the unsatisfactory performance of the few-shot detector, we apply weak data augmentation to both the $N$-way $K$-shot data and the test data, including random resizing and random horizontal flipping. For the hyper-parameters, we set $\lambda_{1}=0.5$ and $\lambda_{2}=0.1$ for all the experiments. We set the thresholds $\delta_{upper}=0.9$ and $\delta_{lower}=0.7$. We optimize the network using Stochastic Gradient Descent (SGD) and set the learning rate to 0.00125. The momentum coefficient of EMA for the teacher network is set to 0.9996.

\begin{table}[t]
\renewcommand\arraystretch{1}
\centering
\caption{Contributions of different components to PS-TTL.}
  \resizebox{0.40\textwidth}{!}{
  \begin{tabular}{c c c| c c c}
    \toprule
    \multirow{2}*{\shortstack{$L_{sup}$}}& \multirow{2}*{\shortstack{$L_{unsup}$}}&  
    \multirow{2}*{\shortstack{$L_{KL}$}}& \multicolumn{3}{c}{nAP50} \\
    & & & 1-shot & 2-shot & 3-shot \\
    \midrule
    &&& 55.4 & 62.1 & 65.0\\
    $\checkmark$&&& 54.3&  61.5& 63.2\\
    &$\checkmark$&& 56.1&  63.4& 65.7\\
    $\checkmark$&$\checkmark$&& 57.0 & 63.8 & 65.4\\
    $\checkmark$&$\checkmark$&$\checkmark$& \textbf{58.4}& \textbf{65.7} & \textbf{67.9} \\
    \bottomrule
  \end{tabular}
  }
  \label{tab:components}
\end{table}

\begin{table}[!t]
\renewcommand\arraystretch{1}
\centering
\caption{Ablation on threshold selection.}
  \resizebox{0.37\textwidth}{!}{
  \begin{tabular}{c c| c c c}
    \toprule
    \multirow{2}*{\shortstack{$\delta_{upper}$}}& \multirow{2}*{\shortstack{$\delta_{lower}$}}&  \multicolumn{3}{c}{nAP50} \\
    & & 1-shot & 2-shot & 3-shot \\
    \midrule
    $-$&$-$& 55.4 & 62.1 & 65.0\\
    0.95&$-$& 55.5 & 61.6 & 64.2\\
    0.90&$-$& \textbf{57.0} & \textbf{63.8} & 65.4\\
    0.85&$-$& 56.8 & 63.4 & \textbf{65.6}\\
    \midrule
    0.90&0.8&  57.1&  65.2& 67.4\\
    0.90&0.7&  \textbf{58.4}&  \textbf{65.7}& \textbf{67.9}\\
    0.90&0.6&  57.5&  65.2& 67.5\\
    \bottomrule
  \end{tabular}
  }
  \label{tab:threshold}
  \vspace{-0.4cm}
\end{table}
 
\subsection{Main Results}
\label{sec:main_results}
 \textbf{PASCAL VOC.} Experimental results on the PASCAL VOC dataset are shown in Table \ref{tab:voc}. We use DeFRCN as our baseline, which incorporates an additional Prototypical Calibration Block (PCB) to refine predictions. However, we find that the $N$-way $K$-shot data utilized by PCB may not align with that used at the novel fine-tuning stage. Therefore, we exclude PCB and present our re-implementation results DeFRCN* in Tables \ref{tab:voc} and \ref{tab:coco}. It can be observed that our method achieves decent improvements across various splits and different shots on the PASCAL VOC benchmark. Our method outperforms HallucFsDet~\cite{zhang2021hallucination} and LVC~\cite{kaul2022label}, which represent synthetic novel class data and semi-supervised learning on base data, respectively. Meanwhile, we find that the performance gain obtained from TTL becomes more significant as the shot number decreases, especially in the 1-shot scenario.

 \textbf{MS COCO.} Table \ref{tab:coco} shows the detection results on MS COCO. The MS COCO dataset conveys more categories, and typically, a single image contains multiple instances. few-shot detectors generally do not perform well on MS COCO due to these factors, which also undermine the performance of our method. However, we observe that our method still achieves a significant improvement compared to the baseline, especially in the mAP75 metric. There is a 5.0\% improvement in AP75 at 10 shots and a 9.2\% improvement in AP75 at 30 shots. LVC \cite{kaul2022label} demonstrates a noticeable improvement on the MS COCO dataset, because the base data in the MS COCO benchmark include a large number of implicit novel instances. However, an issue arises from its setting of few-shot detection, which does not match the real-world scenario. In contrast, under the TTL setting, we only have 5,000 images available for mining implicit novel instances.

\begin{table}[!t]
\renewcommand\arraystretch{1}
\centering
\caption{Ablation on class prototype update.}
  \resizebox{0.35\textwidth}{!}{
  \begin{tabular}{c |c c c c}
    \toprule
    \multirow{2}*{\shortstack{Update Methods}}&  \multicolumn{3}{c}{nAP50} \\
     & 1-shot & 2-shot & 3-shot \\
    \midrule
    Static& 57.5 & 65.1 & 67.7 \\
    Dynamic& \textbf{58.4} & \textbf{65.7} & \textbf{67.9}\\
    \bottomrule
  \end{tabular}
  }
  \label{tab:prototype}
\end{table}

\begin{table}[!t]
\renewcommand\arraystretch{1}
\centering
\caption{Ablation on different data augmentation techniques.}
  \resizebox{0.45\textwidth}{!}{
  \begin{tabular}{c c| c c c}
    \toprule
    \multirow{2}*{\shortstack{Student Aug.}}& \multirow{2}*{\shortstack{Teacher Aug.}}&  \multicolumn{3}{c}{nAP50} \\
    & & 1-shot & 2-shot & 3-shot \\
    \midrule
    Strong&Weak& 56.9 & 64.9 & 66.6\\
    Weak&Weak& \textbf{58.4} & \textbf{65.7} & \textbf{67.9}\\
    \bottomrule
  \end{tabular}
  }
  \label{tab:augmentation}
  \vspace{-0.4cm}
\end{table}

\begin{figure*}[t]
  \centering
  \includegraphics[width=0.95\textwidth]{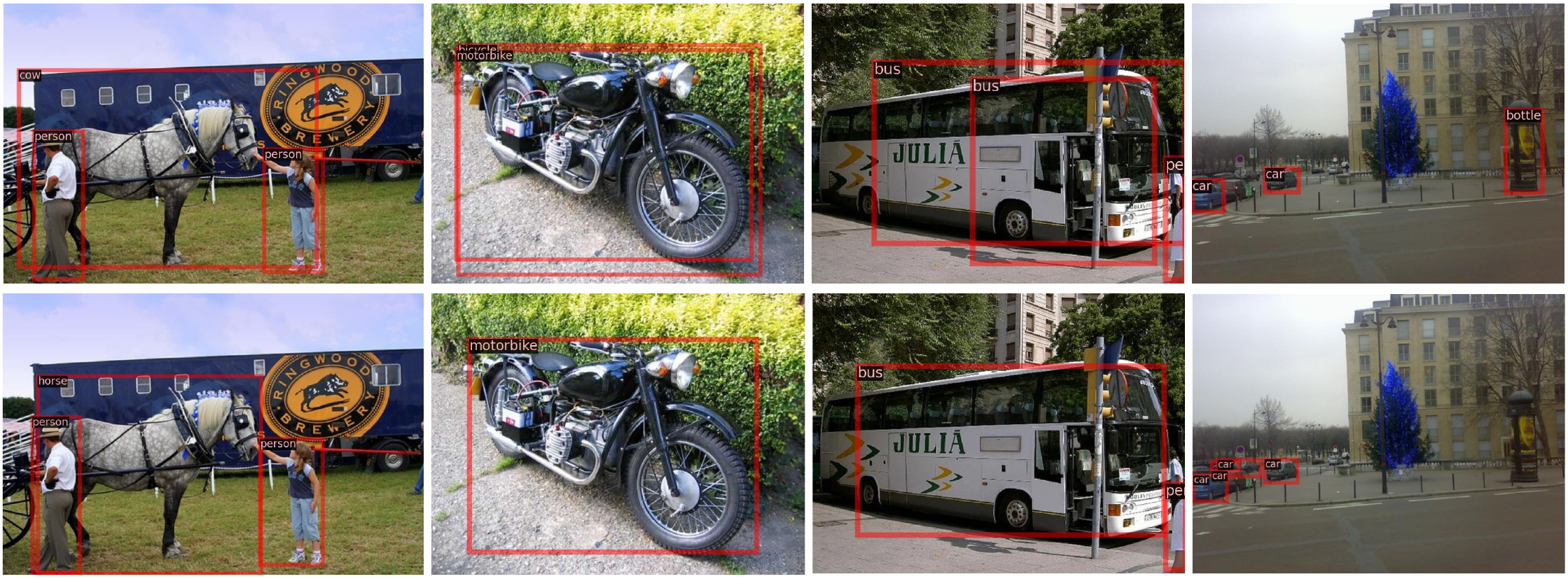}
  \vspace{-0.2cm}
  \caption{Qualitative visualization comparison on PASCAL VOC. The top and bottom lines respectively show the results of DeFRCN and our PS-TTL.}
  \vspace{-0.2cm}
  \label{fig:visual}
\end{figure*}

\subsection{Ablation Studies}
\label{sec:ablation}
 Ablation experiments are carried out on novel split 1 of the PASCAL VOC benchmark to highlight the necessity of the main components of the proposed method.

\subsubsection{Effectiveness of each component.}
We conduct a detailed ablation study on each component of our method, as shown in Table \ref{tab:components}. The first row presents our baseline, \emph{i.e.} DeFRCN. Initially, we attempt to solely utilize $N$-way $K$-shot data for supervised learning during testing but find that the model tends to overfit to these data, resulting in decreased performance. In the third row, we only fine-tune the model using high-quality pseudo-labels at the test phase, yielding the results superior to those of the baseline model. To further enhance the performance in low-sample scenarios, we combine $N$-way $K$-shot data with pseudo-labels for fine-tuning. It is observed that except for the 3-shot setting, the model achieves extra gains in other cases, suggesting that it effectively prevents the accumulation of biases. Finally, by introducing $L_{KL}$, \emph{i.e.}, employing the PS strategy in testing, the model significantly improves its performance across various sample sizes. This also indicates that our proposed method utilizes pseudo-labels in a more efficient way.

 \subsubsection{Upper and lower threshold setting.}
Threshold selection is crucial in pseudo-labeling methods, and we thus conduct ablation experiments on the thresholds, as shown in Table \ref{tab:threshold}. Firstly, we use a large threshold $\delta_{upper}$ to filter high-quality pseudo-labels as hard labels for training the student network. To determine the appropriate value of $\delta_{upper}$, we perform standard self-training on test data without using soft-labels. From Table \ref{tab:threshold}, it can be observed that a larger threshold may incur the problem that only a few pseudo-labels are available as hard labels. This often leads to many foreground objects being mistakenly classified as background, degrading the detection performance of the model. Conversely, a smaller threshold tends to introduce excessive noisy labels, which also negatively affects the performance. By comparing the results from rows 1 to 4, we set $\delta_{upper}=0.9$. Next, we conduct experiments on the threshold $\delta_{lower}$, where prediction boxes with confidence scores between $\delta_{upper}$ and $\delta_{lower}$ are considered as implicit foreground predictions and assigned soft-labels. Similarly, setting $\delta_{lower}$ too high may result in only a few implicit foreground predictions available as soft-labels, while setting it too low is prone to suffer many false positives, mistaking background as implicit foreground. From Table \ref{tab:threshold}, we set $\delta_{lower}=0.7$, which helps the model efficiently utilize implicit foreground predictions, especially in extremely low-shot scenarios (\emph{i.e.}, shot=1). Additionally, we notice that when the few-shot detector performs well, it is not sensitive to $\delta_{lower}$. The reason probably lies in that implicit foreground predictions are correctly assigned higher confidence scores, and background is given lower confidence scores.

\subsubsection{Class prototype update.}
As mentioned earlier, we utilize the feature similarity between low-confidence pseudo-labels and class prototypes to generate soft-labels, where well-defined class prototypes can produce more accurate soft-labels for implicit foreground predictions. However, since we initialize class prototypes using $N$-way $K$-shot data, the features of objects in each class often change during TTL, and static prototypes cannot accurately represent their respective classes. We thus propose to dynamically update class prototypes using high-confidence pseudo-labels, aiming to gradually converge the prototypes to their true class distributions during TTL. In Table \ref{tab:prototype}, we compare the results of static class prototypes and dynamic ones across multiple samplings, with the latter showing consistent improvements. We observe that the improvement by dynamically updating class prototypes becomes more pronounced as the number of samples decreases.

\subsubsection{Alternative data augmentation.}
We also validate the data augmentation techniques used for both the student and teacher networks. Generally, in semi-supervised object detection, weak augmentation is applied to input images for the teacher network, while strong augmentation is used for the student network (please refer to \cite{liu2021unbiased} for more details). However, in our case, we find that even with weak data augmentation for the student network, its performance improves. As shown in Table \ref{tab:augmentation}, consistently using weak-weak data augmentation enhances the performance across all the settings. This is because, during TTL, we can only fine-tune on test data for one epoch. Additionally, in the presence of data scarcity, strong data augmentation tends to disrupt the original data distribution, impeding model convergence.

\subsubsection{Qualitative evaluation.}
 We visualize the detection results of 1-shot of PASCAL VOC in Fig. ~\ref{fig:visual}. Our method significantly alleviates the problem of classification confusion between base classes and novel classes. In the first column, DeFRCN misclassifies a base class (horse) as a novel class (cow), and in the second column, it misclassifies a novel class (motorcycle) as a base class (bicycle). Our method addresses this issue through fine-tuning at test-time. In the third column, DeFRCN predicts multiple local regions of a bus (novel class) as the bus category. Although we do not specifically design any loss for regression, the improvement in classification performance does help the model alleviate this issue. Additionally, our method improves the performance of base classes. For example, in column 4 of Fig.~\ref{fig:visual}, DeFRCN incorrectly identifies a newsstand as a bottle and incompetently misses dense cars, both of which are refined by our method.

\section{Conclusion}
 This paper proposes a novel framework for few-shot object detection, namely Prototype-based Soft-labels and Test-Time Learning (PS-TTL). It aims to address the challenge of precisely capturing the real data distribution of novel classes with scarce labeled samples. To this end, we propose a Test-Time Learning (TTL) module to discover novel instances from test data, allowing detectors to learn better representations and classifiers for novel classes. Furthermore, we design a Prototype-based Soft-labels (PS) strategy to unleash the potential of low-quality pseudo-labels, thereby significantly mitigating the constraints posed by few-shot samples. Extensive experiments are conducted on PASCAL VOC and MS COCO, and PS-TTL achieves the state-of-the-art performance, validating its effectiveness.

\begin{acks}
This work is partly supported by the National Natural Science Foundation of China (No. 62022011), the Research Program of State Key Laboratory of Complex and Critical Software Environment, and the Fundamental Research Funds for the Central Universities.
\end{acks}

\bibliographystyle{ACM-Reference-Format}
\bibliography{sample-base}

\clearpage

\setcounter{table}{0}
\setcounter{figure}{0}
\setcounter{section}{0}
\renewcommand\thesection{\Alph{section}}
\renewcommand\thefigure{\Alph{figure}}
\renewcommand\thetable{\Alph{table}}

\noindent \textbf{\Large Supplementary Material} \\

\appendix

This supplementary material provides more experiment results on PS-TTL in Sec.~\ref{sec:A} and more visualization results in Sec.~\ref{sec:B}.

\section{More Experiment Results}
\label{sec:A}

\subsection{One Batch \emph{vs.} One Epoch}

We implement test-time learning by fine-tuning on test data for one epoch, followed by testing on the same test data. Hence, we require a specific storage capacity to accommodate test data. In a more realistic scenario, we need to conduct test-time learning on sequentially streamed test data. When a batch of testing samples arrives, we first use the detector to make predictions. And then, the detector's weights are updated on this batch of testing samples. We compared two testing strategies, One Batch and One Epoch, on novel split 1 of the PASCAL VOC benchmark, as shown in Table \ref{tab:test-strategy}. By comparing row 1 and row 2, we found that under the One Batch testing strategy, test-time learning can still bring stable improvements to the baseline. Although the performance of the One Batch strategy is generally inferior to that of the One Epoch strategy, we believe this is due to the detector not learning enough knowledge from test data at the early stage during testing.

\subsection{The Performance Trend of Test-Time Learning}

\begin{figure}[t]
  \centering
  \includegraphics[width=0.9\linewidth]{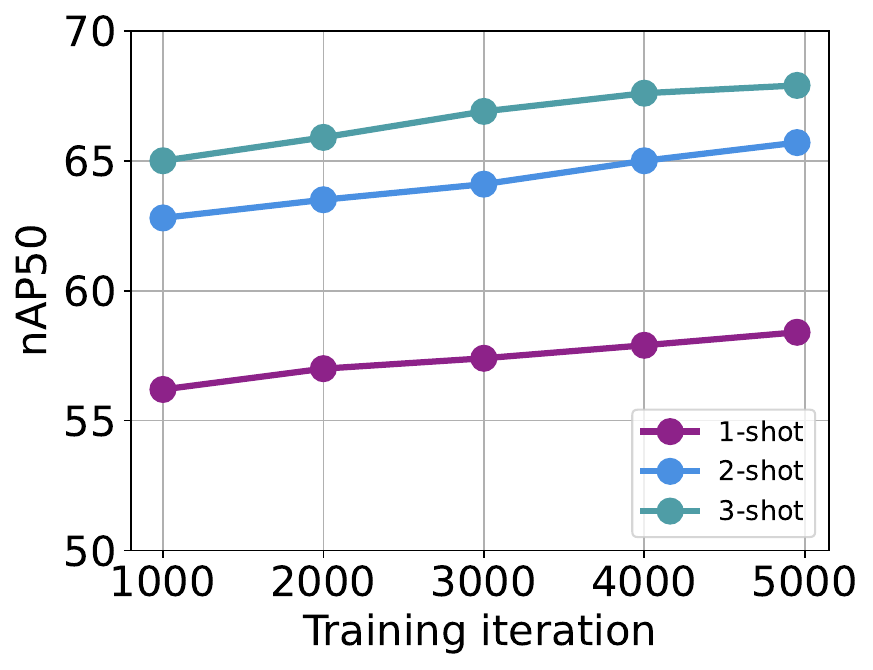}
  \caption{The performance trend of Test-Time Learning in terms of nAP50 (\%) on PASCAL VOC.}
  \label{fig:performance-trend}
\end{figure}

As shown in Fig. \ref{fig:performance-trend}, we plot the performance trend of the detector under different shot settings on novel split 1 of the PASCAL VOC dataset as training iterations progress. As expected, with the increase of training iterations, the performance of the detector improves progressively. It demonstrates the effectiveness of test-time learning. Through test-time learning, we endow the detector with the continuously learning ability. By learning on test data, the detector can better utilize novel instances in test data to capture the data distribution of novel classes. 

\subsection{More Baseline Detectors}

As shown in Tab.~\ref{tab:prototype}, we apply our method to another baseline detector, MFDC \cite{wu2022multi}. The results of the last four rows indicate that our method achieves improvements across different shot settings on novel split 1 of the PASCAL VOC benchmark, demonstrating the effectiveness of PS-TTL.

\subsection{Comparison with Recent SOTA Methods}

DE-ViT\cite{zhang2024detectexamples} is new SOTA few-shot detector integrating DINOv2\cite{oquab2023dinov2} ViT into the framework. It is unfair to directly compare DE-ViT with our method, as DE-ViT takes DINOv2 as the backbone, which means that the few-shot novel classes used for testing may have been seen during DINOv2 pre-training. However, as shown in Tab.~\ref{tab:prototype}, our method with MFDC outperforms DE-ViT in extremely low-shot scenarios (i.e., shot=1/2/3), and reports satisfactory performance at 5/10 shots.

\begin{table}[!t]
\renewcommand\arraystretch{1}
\centering
\caption{Ablation on different test strategies.}
  \resizebox{0.35\textwidth}{!}{
  \begin{tabular}{c |c c c c}
    \toprule
    \multirow{2}*{\shortstack{Test Strategy}}&  \multicolumn{3}{c}{nAP50} \\
     & 1-shot & 2-shot & 3-shot \\
    \midrule
    DeFRCN* &55.4 & 62.1& 65.0 \\
    One Batch& 56.4 & 64.0& 66.4 \\
    One Epoch& {58.4} & {65.7} & {67.9}\\
    \bottomrule
  \end{tabular}
  }
  \label{tab:test-strategy}
\end{table}

\begin{table}[!t]
\centering
\renewcommand{\arraystretch}{1.0}
\caption{Performance (\%) of FSOD methods on PASCAL VOC.}
  \resizebox{0.45\textwidth}{!}{
  \begin{tabular}{c | c c c c c}
    \toprule
    \multirow{2}{*}{\shortstack{Method / Shots}} & \multicolumn{5}{c}{Novel Split 1} \\
    \cmidrule(lr){2-6} 
     & 1-shot & 2-shot & 3-shot & 5-shot & 10-shot\\
    \midrule
    DE-ViT(ViT-S/14)  & 47.5 & 64.5 & 57.0  & 68.5 & 67.3 \\
    DE-ViT(ViT-B/14)  & 56.9 & 61.8 & 68.0  & \textbf{73.9} & \textbf{72.8} \\
    DE-ViT(ViT-L/14)  & 55.4 & 56.1 & 68.1 & 70.9 & 71.9 \\
    DeFRCN* & 55.4 & 62.1 & 65.0  & 68.4 & 67.6 \\
    DeFRCN*+PS-TTL & 58.4 & 65.7 & 67.9  & 69.3 & 68.1 \\
    MFDC* & \underline{62.4} & \underline{67.5} & \underline{68.9} & 71.0 & 71.4 \\
    MFDC*+PS-TTL & \textbf{63.1} & \textbf{69.0}& \textbf{69.1} & \underline{71.3} & \underline{72.4} \\
    \bottomrule
  \end{tabular}
  }
  \label{tab:prototype}
\end{table}

\begin{table*}[t]
\renewcommand\arraystretch{1}
  \centering
  \caption{Performance (\%) of FSOD methods on MS COCO.}
\resizebox{0.6\textwidth}{!}{
  \begin{tabular}{l| c c | c c | c c | c c}
    \toprule
    \multirow{2}{*}{Method}  &  \multicolumn{2}{c|}{1-shot} &  \multicolumn{2}{c|}{2-shot} &  \multicolumn{2}{c|}{3-shot} & \multicolumn{2}{c}{5-shot}\\
    & nAP & nAP75 & nAP & nAP75 & nAP & nAP75 & nAP & nAP75\\
    \midrule
    TFA w/cos~\cite{wang2020frustratingly}&3.4 & 3.8&4.6  & 4.8 &6.6& 6.5& 8.3& 8.0\\
    MPSR~\cite{wu2020multi}&2.3& 2.3& 3.5&3.4& 5.2& 5.1 &6.7& 6.4\\
    QA-FewDet~\cite{han2021query}&4.9 & 4.4 &7.6 &6.2&8.4&7.3&9.7&8.6 \\
    FADI~\cite{cao2021few}&5.7& 6.0 &7.0 &7.0&8.6&8.3&10.1&9.7 \\
    DeFRCN*~\cite{qiao2021defrcn}& 5.5 &5.7 &9.8 & 9.9&12.3&\textbf{12.6}& 14.2& 13.7 \\
    \midrule
    Ours& \textbf{6.0} & \textbf{6.5} & \textbf{10.1} & \textbf{10.3} & \textbf{12.5 }&12.5 &\textbf{14.4} &\textbf{13.8 } 
    \\
    \bottomrule
  \end{tabular}
  }
  \label{tab:coco-low-shot}
\end{table*}

\begin{figure*}[t]
  \centering
  \includegraphics[width=0.85\textwidth]{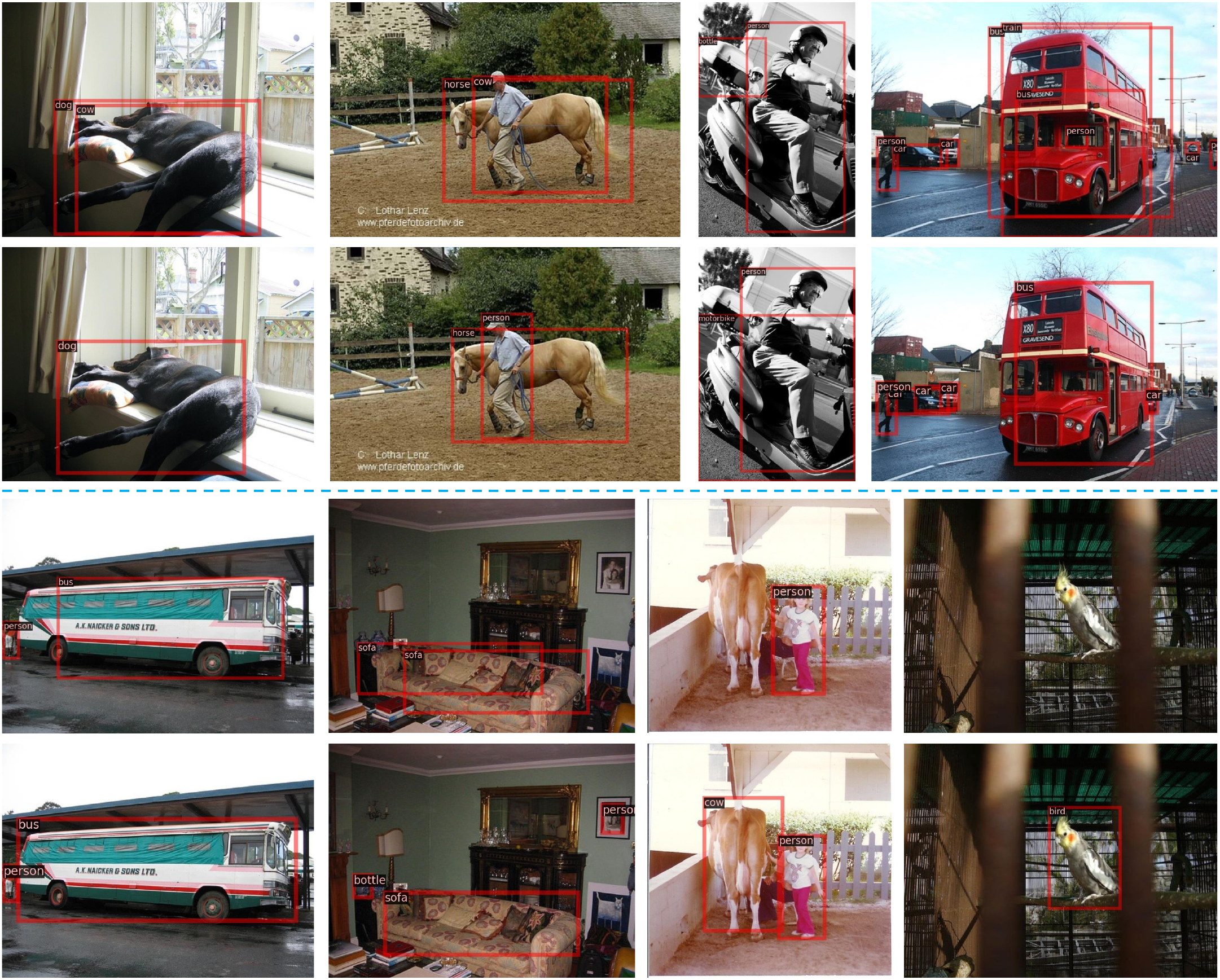}
  \caption{Qualitative visualization comparison on PASCAL VOC. The top and bottom lines in each group respectively show the results of DeFRCN and our PS-TTL.}
  \label{fig:visual2}
  \vspace{-0.1cm}
\end{figure*}

\subsection{Results on MS COCO Under Low-shot Settings}

Table \ref{tab:coco-low-shot} shows detection results on MS COCO under low-shot settings. The MS COCO dataset contains 80 categories, with each image typically containing multiple instances. This leads to a degradation in the performance of few-shot detectors on test data, especially in low-shot settings, which hinders the ability to conduct test-time learning. 
However, our method consistently improves across various low-shot settings, especially in extremely low-sample scenarios, demonstrating notable enhancements. For example, in the 1-shot scenario, our method improved the mAP on novel classes by {9\%} compared to DeFRCN.

\section{More Visualization Results}
\label{sec:B}

We visualize more detection results of 1-shot on PASCAL VOC in Fig. \ref{fig:visual2}. Our method can alleviate the misclassification issue between base and novel classes, as shown in the top group of Fig. \ref{fig:visual2}. DeFRCN misclassifies base class dogs and horses as novel class cows, novel class motorbikes as base class bottles, and novel class buses as base class trains. Although our method is not optimized for the regression branch, as more novel class instances are observed, our method can improve the regression performance of novel classes, as shown in the first two columns of the bottom group in Fig. \ref{fig:visual2}. Furthermore, in the construction of base data for FSOD, there are numerous unlabeled novel instances in base data. This may result in some novel instances being misclassified as background. Our method continuously learns on test data, which helps mitigate this issue. As shown in the column 3 and column 4 of the bottom group in Fig. \ref{fig:visual2}, our method can prevent the omission of novel classes, such as cows and birds.

\end{document}